\documentclass[draft]{agujournal2019}
\usepackage{url} 
\usepackage{lineno}
\usepackage[finalnew]{trackchanges} 
\usepackage{soul}

\usepackage{amsmath}
\usepackage{amsfonts}
\usepackage{bm}
\usepackage{graphicx}
\usepackage{etoolbox}
\usepackage[makeroom]{cancel}
\usepackage{amssymb,mathtools}

\newcommand*\linenomathpatch[1]{%
  \cspreto{#1}{\linenomath}%
  \cspreto{#1*}{\linenomath}%
  \csappto{end#1}{\endlinenomath}%
  \csappto{end#1*}{\endlinenomath}%
}

\linenomathpatch{equation}
\linenomathpatch{align}

\draftfalse

\journalname{Water Resources Research}

\begin{document}

%
%


\title{How to select an objective function using information theory}

%
%




\authors{Timothy O. Hodson\affil{1}, Thomas M. Over\affil{2}, Tyler J. Smith\affil{3}, and Lucy M. Marshall\affil{4}}


\affiliation{1}{U.S. Geological Survey Water Resources Mission Area, Urbana, Illinois}
\affiliation{2}{U.S. Geological Survey Central Midwest Water Science Center, Urbana, Illinois}
\affiliation{2}{Clarkson University, Potsdam, New York}
\affiliation{3}{Macquarie University, Sydney}




\correspondingauthor{Timothy Hodson}{thodson@usgs.gov}



\begin{keypoints}
  \item A basic problem in modeling is the choice of objective function (or performance metric)
  \item According to information theory, the ``best'' objective function minimizes information loss, which we evaluate using the AIC
  \item Like friction or inefficiency in a system, information loss incurs additional cost however the model is used
\end{keypoints}

%
%

%
%


\begin{abstract}
  In machine learning or scientific computing, model performance is measured with an objective function.
  But why choose one objective over another?
  According to the information-theoretic paradigm, the ``best'' objective function is whichever minimizes information loss.
  To evaluate different objectives, transform them into likelihoods.
  The ratios of these likelihoods represent how strongly we should prefer one objective versus another,
  and the log of that ratio represents the relative information loss (or gain) from one objective to another.
  In plain terms, minimizing information loss is equivalent to minimizing uncertainty,
  as well as maximizing probability and general utility.
  We argue that this paradigm is well-suited to models
  that have many uses and no definite utility 
  like the complex Earth system models used to understand the effects of climate change.
  Furthermore, the benefits of ``maximizing information and general utility'' extend beyond model accuracy to other important
  considerations including how efficiently the model calibrates, how well it generalizes, and how well it compresses data.
\end{abstract}

%
%

%


%
%
%
%

\section{Introduction}
Science tests competing theories or models by evaluating the similarity of their predictions against observational experience,
favoring those that best fit the evidence.
Thus, how scientists measure similarity fundamentally determines what they learn.
In machine learning and scientific computing, an analogous process occurs when calibrating or evaluating a model.
In that case, the model is ``tuned'' according to a similarity metric,
known as an objective or loss function.
A classic example is the mean squared error,
which is the optimal measure of similarity when errors are normally distributed and independent and identically distributed (\textit{iid}). 
However, for many problems, the true error distribution is complex or unknown.
Rather than simply assuming some de facto objective function,
we argue that information theory should guide that choice.

The debate series led by \citeA{Kumar_2020} posits that information theory provides a new paradigm for Earth science.
Here we give a basic, yet motivating, demonstration by using information theory to help solve a fundamental question: 
How should we select an objective function?
\citeA{Weijs_2010} provide an excellent summary of the two main paradigms for doing so,
and their conclusions lead naturally to this paper.
The first paradigm is more related to engineering in that it tries to maximize a particular value or utility of a model.
Take the example of flood forecasting.
In this value-based paradigm, the objective might be to minimize the expected damage caused by erroneous flood predictions.
In this sense, the paradigm is subjective because it prioritizes a specific set of values.
However, many models do not have a defined purpose and might serve conflicting interests.

Consider a cosmological model or a global climate model.
Such models have no definite or singular utility,
yet we consider them to be of immense societal importance and value.
Instead, we can adopt the information-theoretic paradigm, and try to minimize the model's uncertainty in a general sense.
In practice, this is done by minimizing the information loss \cite<sometimes referred to as K-L information loss after>[]{Kullback_1951}
measured against a set of reference observations.
The ultimate objective in this paradigm is to minimize general uncertainty (and maximize information).
However, this can also be adapted to the value-based paradigm
to minimize uncertainty in a particular application of a model.
Case in point, we will demonstrate how to select the best general performance
metric for a particular model.

So why not always adopt the information-theoretic approach?
Invariably there are practical reasons to deviate:
Information is exceedingly difficult to measure.
In practice, we always approximate it,
but that hardly detracts from its utility as a guiding scientific framework.
Even state-of-the-art scientific models rely heavily on ``folk wisdom'' 
\cite<i.e., substantial amount of model-develop\-ment lore that developed ad hoc rather than following some overarching theory;>{Wolfram_2023}.
Some of that wisdom is probably well-founded; some is not;
and information theory---as well as the related fields of probability theory and algorithmic information theory---helps distinguish between the two.
For example, much of the ``best practice'' and ``folk wisdom'' that modelers understand intuitively have foundations in information theory,
including using cross-validation to mitigate over\-fitting \cite{Hastie_2009_ESLII}
or the gradual replacement of mean squared error, which was popular within machine learning in the 1980s and 1990s,
with cross-entropy losses and the principle of maximum likelihood \cite{Goodfellow_2016}.

Before proceeding, two definitions are helpful.
The ``model'' is the knowledge or theory that explains or captures information shared among some set of variables.
All models are approximations: they explain some things but not everything.
The ``uncertainty'' represents the information that cannot be explained by the model.
Our representation of uncertainty is also a model---everything we know is---but the distinction is helpful
in the discussion that follows.
In the information-theoretic paradigm, our goal is to minimize uncertainty,
so our objective function should represent the model's uncertainty as accurately as possible \cite<>[]{Weijs_2020}.
Given a particular dataset and model, we can evaluate how well an objective represents the uncertainty
using standard statistical methods.
This paper focuses on one in particular, the Akaike Information Criterion \cite<AIC;>{Akaike_1974},
which is easy to compute and very accurate for evaluating objective functions.
After reviewing some of the theoretical background,
we demonstrate the basic methodology of using AIC to select an objective
function for a streamflow model.

\section{The Experiment}
In the classic modeling experiment, a model is varied (or ``tuned'') while the test data and objective are held fixed.
To select the ``best'' model, choose whichever model optimizes the objective function computed on the test data.
If mean squared error (MSE) is the objective, compute the MSE between the test data and the model predictions,
then select the model with the lowest MSE.
To select the ``best'' objective, the experiment is similar but flipped
such that the objective is varied while the model and data are held fixed.
Now, select the objective indicating the greatest similarity between the data and the model.
Different objective functions have different scales, so they are normalized
such that each integrates to one, thereby representing them as probability distributions.
Intuitively, we cannot directly compare an error of $1 ^{\circ}\text{C}$ to an error of $1 ^{\circ}\text{F}$ to an error of $1\%$ because they are on different scales,
so we normalize them.

The normalized form of MSE is the normal distribution \cite{Hodson_2022}.
When used to evaluate model fit, we refer to that normalized distribution as the \textit{likelihood function}
and its output the \textit{likelihood}.
To select among objectives (or any model), compare their likelihoods and favor the most likely \cite<Maximum Likelihood Principle;>{Fisher_1922}.
Taking the natural logarithm of the likelihood, denoted as $\ell$,
does not change the model ranks
but simplifies the math by converting products to sums:
likelihoods multiply, so log-likelihoods add.
Besides being easier to compute, $\ell$ also represents the information in the model's error (up to a constant),
which is also equivalent to the model's uncertainty.
Technically, that equivalence is only asymptotic, but it is still useful in the same sense that MSE is.

Thus far, the problem is framed in terms of probability theory,
but information theory gives an equally valid interpretation.
The goal of the former is to maximize probability,
whereas, the goal of the latter is to maximize information, which are equivalent \cite{Cover_2006}.
The formal connections between concepts like information, uncertainty, probability, similarity, objective functions, likelihoods,
and data compression are reviewed in the next section.
Readers may also skip to the results and implications (Section~\ref{section_demonstration}), and then return here for theory and explanations.

\section{Uncertainty to Information} \label{section_information}
\label{sec.uncertainty-to-information}
\textit{To maximize the information in the model, select the most likely objective function or whichever represents the error in the fewest bits}.
The explanation follows and summarizes elements of \citeA{Cover_2006} and \citeA{Burnham_2002_MMI2}.

Three fundamental concepts in information theory are
(1) the entropy $H(D)$, which is the expected information in each new observation of the data $D$;
(2) the conditional entropy $H(D|M)$,
which is the additional information needed to represent $D$ after encoding it with some model $M$
(think of it as the information in the model error or ``the model's uncertainty'');
(3) and the difference between terms 1 and 2, known as mutual information,
\begin{equation}
  I(D; M) = H(D) - H(D | M) \text{,}
\end{equation}
which measures how much information $M$ encodes about $D$.
Throughout the paper, the concepts of ``entropy'' and ``uncertainty'' are interchangeable \cite<e.g.>{Shannon_1948}.
In practice, we use a model's errors to evaluate its ``conditional entropy,''
which is just a formal term for a model's ``uncertainty.''

When comparing models against the same data, $H(D)$ is constant $C$
so we only need to compute $H(D|M)$ to assess relative differences in $I$.
Now, the connection to probability theory:
The entropy of a random variable $X$ with probability density function $f(x)$ is defined as
\begin{equation}
  H(X) = - \int f(x) \log f(x) dx = E_X[-\log(f(X))] \text{.}
  \label{eq.entropy}
\end{equation}
Substituting the likelihood for $f(x)$ and taking the limit as the number of observations $n$ goes to infinity,
the log-likelihood $\ell$ equals the negative conditional entropy and also the mutual information up to a constant ($H(D)$),
\begin{equation}
  I_C = - H(D|M) = \lim_{n \to \infty} \frac{\ell}{n \ln(2)} \label{eq.equivalence} \text{,}
\end{equation}
where the natural logarithm gives units of nats
and dividing by $\ln(2)$ converts the result to bits.
In other words, sampling the error distribution and computing $\ell$ performs a Monte Carlo integration of the log-likelihood (Equation~\ref{eq.entropy}),
which yields the conditional entropy.
For finite $n$, the average $\ell$ gives an unbiased estimate 
of $I_C$ unless the data used to estimate $\ell$ were also used to fit the model parameters, which causes ``overfitting'' \cite{Akaike_1974}.
It is evident from Equation~\ref{eq.equivalence} that minimizing the entity in the second term maximizes that in the first and third.
Therefore, to maximize the information in the model, select the most likely objective function or whichever represents the error in the fewest bits.

To give some additional intuition, we will explain this idea another way.
In practice, we assume a model for $\ell$ (like the normal distribution),
which we substitute for the true error distribution $\ell_o$, which is not known.
In doing so, we pay a penalty in terms of information loss $K$,
\remove{such that our estimate of $\ell$ can be split into two components,}
\begin{equation}
  \lim_{n \to \infty} \frac{\ell}{n \ln(2)} = \lim_{n \to \infty} \frac{\ell_o}{n \ln(2)} - K(\ell_o, \ell) \text{,}
  \label{eq.kl-divergence}
\end{equation}
the true likelihood $\ell_o$ and a penalty $K(\ell_o, \ell)$,
which is the Kullback-Leibler divergence or relative entropy between our model of $\ell$ (represented by the objective function)
and the true likelihood $\ell_o$.

Essentially, $K$ measures the information loss, in bits, we incur for assuming the wrong error distribution with our objective function.
As we discuss throughout the paper, whatever the intended use of our model, we will pay that penalty in some form.
Later on, we will demonstrate how to determine which objective pays the smallest such penalty.

Because of its generality, information theory bridges all scientific disciplines \cite[pp. 1--5]{Cover_2006, Kumar_2020},
which creates several interest\-ing taut\-ologies---
maximizing information is equivalent to minimizing uncertainty, is equivalent to maximizing probability, is equivalent to minimizing code length, is equivalent to maximizing data compression \cite{Grunwald_2007}---
that are repeated and explored throughout this paper.
In that broad sense, the information-theoretic paradigm motivates a wide range of well-known statistical methods including 
generalized linear models, Bayesian model selection, and ``universal'' likelihood or divergence estimators \cite<e.g.,>{Perez_Cruz_2008, Wang_2009, Vrugt_2022_universal}.
We use the term ``universal'' somewhat informally to refer to methods that are not bound to a specific parametric likelihood function.
We will not review them here, but they are, in theory, all justified and evaluated using information theory.

\section{Objectives to Log Likelihoods}
\begin{figure}
  \centering
  \includegraphics[width=8.3cm]{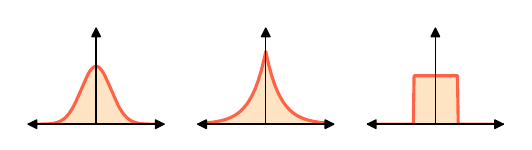}
  \caption{Normal, Laplace, and uniform error distributions, respectively. }
  \label{figure1}
\end{figure}

Given a large dataset,
we can estimate the conditional entropy $H$ of different objectives using the maximum likelihood estimate of their log-likelihoods
(Equation \ref{eq.equivalence}).
The first, and arguably de facto, objective is MSE, 
which corresponds to the log-likelihood of the normal distribution (Figure \ref{figure1}), which is given by
\begin{equation}
  \ell_2 = -n \ln \sigma  - \frac{n}{2} \ln(2\pi) - \frac{1}{2\sigma^2} \sum_{i=1}^n (y_i - \hat y_i)^2 \text{,}
\end{equation}
where $y_i$ are the observations, $\hat y_i$ are the model predictions, and $\sigma$ is standard deviation of the error.
The final term of the log-likelihood represents the squared error in the objective, and the remaining terms normalize the result
($\ell_2$ is a mnemonic reference to MSE, which is related to the L2 norm; \ref{appendix_nse}).
Another common objective is the mean absolute error (MAE),
which corresponds to the log-likelihood of the Laplace distribution (Figure \ref{figure1}), which is given by
\begin{equation}
  \ell_1 = -n \ln(2b)  - \frac{1}{b} \sum_{i=1}^n | y_i - \hat y_i| \text{,}
\end{equation}
where $b$ is the mean absolute error
($\ell_1$ refers to the L1 norm).

Likelihoods for a variety of other objective functions are obtained by changing variables.
For example, the mean squared log error (MSLE), which corresponds to the lognormal log-likelihood $\ell_3$,
is obtained from $\ell_2$ by changing variables
\begin{equation}
  \ell_3 = \ell_2(v(y)) + \ln|v'(y)| \text{,} \label{eq.change-of-variables}
\end{equation}
where $v$, the natural log in this case,
can be substituted with other functions to obtain log-likelihood ``equivalents'' for
normalized squared error \cite<NSE;>{Nash_1970}, mean squared percent error (MSPE), 
as well as their Laplace equivalents.
``Equivalence'' here means they are equivalent in a maximum likelihood sense that
optimizing on $\ell$ yields the same parameter set as optimizing on the objective function
\cite<but not to be confused with the actual $\ell$ of NSE,
which would be used to estimate a confidence interval; e.g.,>{Vrugt_2022}.
The classic NSE and MSE are only equivalent in the former sense (\ref{appendix_nse}),
meaning they are redundant as objectives,
so the demonstration uses a common variant of NSE.
Derivations and additional explanations of the log-likelihoods are given in \ref{appendix_loglikelihood}.

Likelihoods can also be combined as mixtures to represent different states.
For example, streamflow can have zero or negative values---as in reverse flow---for which $\ell_3$ is undefined.
The ``zero flow'' state is handled by mixing
a binomial distribution for the zeros, with a lognormal for the non-zeros \cite{Smith_2010}.
Taking $n_1$ as the number of correctly predicted zero flows, and
$n_2$ as the number of incorrect zero flows,
the probability of correctly predicting the flow state among $n_1$ and $n_2$ is estimated as $\rho = n_1/(n_1 + n_2)$.
The corresponding binomial log-likelihood is
\begin{equation}
  \ell_0 = n_1 \ln \rho + n_2 \ln (1-\rho) \text{,}
\end{equation}
and the mixture is
\begin{equation}
  \ell_4 = \ell_0 |_{n_1 + n_2} + \ell_3|_{n_3} \text{,} \label{eq.zero_inflated_likelihood}
\end{equation}
where $\ell_0$ is evaluated over the zeros ($n_1 + n_2$), and $\ell_3$ is evaluated over the remaining observations ($n_3$).
This type of binomial mixture is sometimes called a ``zero-inflated'' distribution for the way it inflates the probability of zeros.
Although we believe mixture distributions are useful in practice, \ref{appendix_limitations} discusses an important and unresolved problem with them.

\section{Overfitting Bias} \label{section_aic}
The convergence of log-likelihood, $\ell$, to conditional entropy ($H$, Equation \ref{eq.equivalence})
only holds when $\ell$ is evaluated ``out of sample,'' meaning the data used to estimate $\ell$ were not also
used to estimate the model parameters, which include the parameters of the likelihood itself (e.g., $\sigma$ in $\ell_2$).
Otherwise, our estimate of $\ell$ will be biased, which is known as overfitting \cite[pp. 60--64]{Burnham_2002_MMI2}.
Various approaches to correcting that bias give rise to well-known information criteria
like AIC, BIC, and others \cite{Akaike_1974, Schwarz_1978}.

Information criteria are difficult or impossible to compute for certain classes of models,
notably deep neural networks \cite{Watanabe_2010},
so in practice, cross-validation is widely used to estimate the unbiased log-likelihood.
But cross-validation is also costly: Data are either omitted or else multiple calibrations are performed on different subsets.
When selecting an objective, however, cross-validation is usually unnecessary.
Objective functions tend to have relatively few parameters,
so overfitting is negligible,
and their parameters are often independent, which is assumed by many information criteria.
Even for dependent parameters, information criteria give a good approximation when the parameters are few enough.

Consider AIC,
\begin{equation}
  \text{AIC} = -\ell + k \text{,}
\end{equation}
where $k$ is the number of independent parameters
\cite<>[note that historically, the formula includes an arbitrary factor of 2,
  which we omit throughout.]{Akaike_1974}.
The first term is the log-likelihood, and the second is the expected overfitting bias.
Note that $\ell$ is $\mathcal{O}(n)$,
such that the relative magnitude of the bias term diminishes as $n$ increases.

Now imagine two experimental scenarios.
In the first, the model structure is fixed while the objective is varied.
The model parameters are independent of the objective parameters,
so we can separate their contribution to overfitting
\begin{equation}
  \text{AIC} = -\ell + k_{1} + k_{2} \text{,}
\end{equation}
where $k_{1}$ and $k_{2}$ are the number of parameters in the model and objective, respectively.
In the experiment, $k_{1}$ is fixed,
so only the parameters in the objective ($k_{2}$) influence our selection.
This means that if our only goal is to select an objective,
we can ignore the parameters in the model entirely.
Alternatively, the experiment might simultaneously vary the model structure and the objective
to determine which combination performs best.
In this case, we could cross-validate to estimate the model parameters ($k_{1}$),
then compute AIC on the holdout data and penalize by $k_{2}$ as in the previous experiment.
For many practical problems with large datasets, the effect of overfitting $k_2$ may be negligible;
nevertheless, it is easy to correct using AIC.

\section{Akaike Weights}
Information criteria are evaluated on a relative scale,
so they are routinely reported as differences
\begin{equation}
  \Delta_i = \text{AIC}_i - \min(\text{AIC}) \text{,}
\end{equation}
such that the ``best'' model in the set has $\Delta_i \equiv 0$.
Here and throughout the paper, ``best'' is always in the information-theoretic sense
of minimizing the information loss.

The ``weight of evidence for'' or ``probability that'' a particular model is ``best'' in the set is then
\begin{equation}
  w_i = \frac{ \exp(-\Delta_i)}{ \sum^{m}_{i=1} \exp(-\Delta_i) } \text{.}
\end{equation}

The exact formula for these so-called Akaike weights varies from source to source depending on the base of the logarithm,
whether AIC includes an arbitrary factor of 2, or whether they use different information criteria \ref{appendix_bayes}.
The point to remember is that $\Delta_i$ measures relative entropy between two models,
such that the Akaike weights provide a means of model averaging,
which can reduce model selection bias when multiple models are probable
\cite[p. 150]{Burnham_2002_MMI2}.
If, however, one model dominates, this sort of averaging has little effect.

One final assumption worth mentioning is that AIC assumes that a true model does not exist, which is reasonable for many scientific problems.
In contrast, BIC assumes that the true model is included in the experimental set $\{1, 2, \dots, m\}$ \cite[pp. 284--289]{Burnham_2002_MMI2}.

\section{Benchmark Demonstration} \label{section_demonstration}
To demonstrate the basic process of selecting an objective,
we compute entropies (uncertainties) and Akaike weights (probabilities) for several objective functions
with a simple streamflow model.
In this experiment, we are not optimizing the model per se but rather
using information theory to select the ``best'' metric of predictive performance,
in the sense that it minimizes information loss and uncertainty.
We will discuss why this is useful later on.

The model predicts flow by scaling the nearest concurrent observation by the ratio of the two drainage areas;
so when predicting flow in a large river using observations from a smaller one, it scales up the observations
accordingly.
The test data are roughly 14 million daily streamflow observations taken from 1,385 streamgages in the conterminous United States \cite{Russell_2020}.
As streamflow can be zero or negative (reverse flow), which is undefined for some objective functions,
flows below 0.0028 m$^3$ s$^{-1}$ (0.01 ft$^3$ s$^{-1}$) were thresholded and treated as the ``zero-flow'' state in the comparison.
Different thresholds may yield slightly different results,
particularly among logged objectives because of their sensitivity near zero
(Strictly, the zero state should be handled using zero-inflated objectives,
but we recognize there may be practical reasons for preferring a more traditional objective).

Table~\ref{table1} gives the conditional entropies ($\hat H$) of each objective measured in bits.
\textit{The best objective represents the error in the fewest bits} (Section~\ref{sec.uncertainty-to-information}).
Larger values of $\hat H$ indicate greater divergence between the empirical error distribution and
the likelihood associated with that objective,
which incurs greater information loss ($K$ in Equation~\ref{eq.kl-divergence}).
Here, the empirical error distribution diverges from the normal distribution,
which is why MSE and NSE have greater entropies (more bits) than MAE, MSLE, MARE, etc, which all assume heavier tails.
In this particular experiment, zero-mean absolute log error (ZMALE) performed best,
with an entropy of 6.95 bits;
whereas, MSE was 11.62 bits, and NSE, the de facto objective in hydrologic modeling, was 11.20 bits.
The magnitude of the entropies is unimportant here, only their differences.
The lower conditional entropy and, therefore, better performance of
objectives that log transform the data makes intuitive sense because hydrologic models typically make greater errors at greater streamflows;
logging reframes the error into proportional terms rather than absolute ones.

\begin{table}
  \caption{Conditional entropies $\hat H$ and Akaike weights of ten objective functions evaluated against the test data and model.}
  \begin{tabular}{llcrrr}
    \hline
    Objective & Description                     & $k$ & $\hat H$ in bits & Weight & Rank \\
    \hline
    MSPE      & mean squared percent error      & 1   & 23.54            & 0.00   & 10   \\
    U         & uniformly distributed error     & 1   & 18.17            & 0.00   & 9    \\
    MSE       & mean squared error              & 1   & 11.62            & 0.01   & 8    \\
    NSE       & normalized squared error*       & 1   & 11.20            & 0.01   & 7    \\
    MAE       & mean absolute error             & 1   & 9.49             & 0.04   & 6    \\
    MSLE      & mean squared log error**        & 1   & 7.47             & 0.15   & 5    \\
    MARE      & mean absolute square root error & 1   & 7.34             & 0.17   & 4    \\
    ZMSLE     & zero-inflated MSLE              & 2   & 7.18             & 0.19   & 3    \\
    MALE      & mean absolute log error**       & 1   & 7.04             & 0.21   & 2    \\
    ZMALE     & zero-inflated MALE              & 2   & 6.95             & 0.22   & 1    \\
    \hline
    \multicolumn{5}{l}{*modified Nash--Sutcliffe efficiency. **Undefined for zero flow but included for context.}
  \end{tabular}
  \label{table1}
\end{table}

In the experiment, the data and model were fixed,
so too were the errors.
All that changed was how we measured the information in the error.
In that sense, the excess bits in the other objective functions are noise \cite<e.g.,>[p. 2]{Rissanen_2007}.
So, MSE measures at least 40 percent noise ($1 - 6.95/11.62$),
and NSE at least 38 percent.
Although this use of the term ``noise'' may seem unusual,
the intuition is straightforward: Noise increases information loss and uncertainty.

To be clear, we are not the first to argue that NSE is a poor performance metric for streamflow prediction \cite<e.g.,>{Clarke_1973, Sorooshian_1980, Gupta_1998, Legates_1999, Clark_2021}
or that some form of maximum likelihood can be used to select a better one \cite<e.g.,>{Sorooshian_1980, Gupta_2003, Schoups_2010, Smith_2015}.
The contribution of this paper is (1) to quantify that intuition in terms of information loss,
(2) provide a formal method for selecting objectives on an information basis,
and (3) provide the theoretical justification for such an approach:
Performance metrics lose information as the empirical error distribution diverges from that implied by their likelihood. 
In this regard, ``universal'' likelihoods, which can fit a range of distributions, should perform better than the simple objectives tested here,
but they should be evaluated in the same way using information theory.

\begin{figure}[!h]
  \centering
  \includegraphics{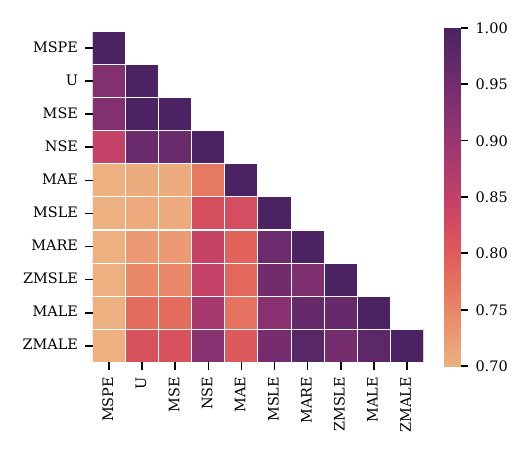}
  \caption{Location-wise correlation in conditional entropy among different objective functions.
    The higher-entropy objectives (MSPE, U, MSE, NSE) are correlated because they all measure errors in absolute terms;
    whereas, the lower-entropy metrics (MSLE, MALE, ZMSLE, ZMALE) all measure error in relative terms.
    The others are between these two extremes.
    For example, MAE measures error in absolute terms but is less sensitive to outliers, which occur
    more frequently as the objective diverges from the error distribution.} \label{figure2}
\end{figure}

Intuitively, it makes sense that different performance metrics or objectives
measure different things and, therefore,
lead to different decisions regarding a model.
Figure~\ref{figure2} shows this in a simplistic sense:
When compared streamgage to streamgage, ``good'' (low information loss) objectives were more correlated with one another
than with ``bad'' ones.
Admittedly, this is a very general observation---different objectives quantify different things---but some general implications follow.

The value-based paradigm selects an objective function for a particular application:
``Each model has a purpose, and we should tailor the model and objective to that purpose.''
But those decisions can have unintended consequences if our application is uncertain.
Moreover, can we ever ensure a model will be used exactly as we expect or foresee every application?
Contrast that with the information-theoretic paradigm to ``select an objective function that maximizes information and general utility.''
What are the concrete benefits of such a general strategy?
This brings us to a fundamental problem with optimization.
When we optimize, we fit the model to a particular dataset,
but our intended application is that the model generalizes to new data.
Generalizability is an inherent benefit of ``maximizing general utility.''
The complexity penalty in AIC is one metric for evaluating how well a model might generalize,
but so is the likelihood function.
Calibrate to a poor likelihood (objective) function, and the model will generalize poorly.
For example, using MSE or NSE with a fat-tailed error distribution will inherently over-fit ``outliers.''
On one hand, this is the intent under much of the prevailing wisdom
---``that we use NSE to emphasize accuracy at high streamflow''---
but information theory explains why this practice is also harmful.

\begin{figure}[!h]
  \centering
  \includegraphics[width=14cm]{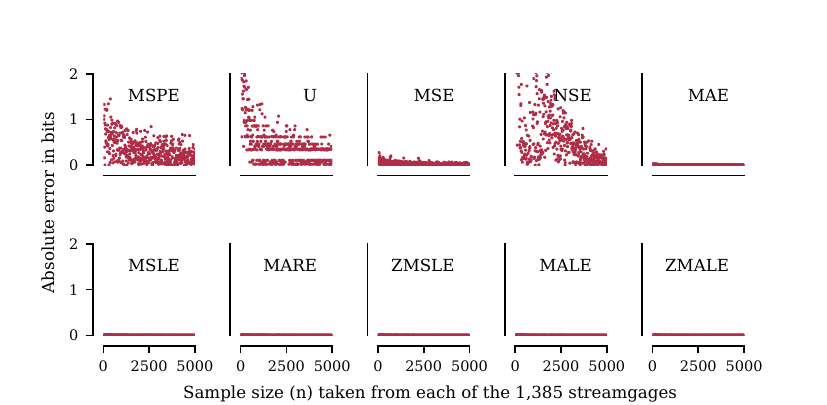}
  \caption{
    The absolute error of the entropies of different objectives versus sample size ($n$),
    where $n$ is the number of observations taken at random from each of the 1,385 streamgages.
    The error for each sample was taken relative to the mean of the five largest samples.
    Lower-entropy objectives generally converge faster, except for NSE,
    which is unusually noisy from uncertainty about $\sigma_o$ (\ref{appendix_nse}).
    Also, note how the objectives on the bottom row appear to have negligible uncertainty relative to those on top.
    These symptoms, and more, are captured in the Akaike weights (Table~\ref{table1}).
  } \label{figure3}
\end{figure}

Our choice of objective also affects how accurately we can estimate model parameters,
including how accurately we can estimate the objective itself.
Figure~\ref{figure3} shows this result as we randomly subsample the data and compute the entropies of the objective functions
for larger and larger subsamples.
Two things are evident.
The uncertainty about the objective decreases as the sample size increases,
but some objectives converge much faster than others.
By no coincidence, those that converge faster also rank better in Table~\ref{table1}.

If our objective is uncertain, it is difficult to determine a good model from a poor one.
Recall that our objective function also represents our model's uncertainty,
so uncertainty about our objective means uncertainty about our uncertainty.
Such a generic statement necessitates a poignant example.
To compute a prediction interval, an expectation, or any quantity from our model,
we must correctly adjust the model's predictions based on its uncertainty.
The likelihood determines the form of that adjustment,
so a poor (lossy) likelihood leads to inaccurate prediction intervals, etc. (\ref{appendix_expectations}).

How we determine a good model from a poor one is the central problem of science, as well as optimization,
and our objective function also affects the rate of convergence during optimization.
An algorithm making noisier parameter updates during each iteration will converge more slowly.
This result is well known in the literature on stochastic optimization \cite{Bottou_2007, Thomas_2020}.
Although most work in this area has focused on general-purpose optimization methods like stochastic gradient descent,
the same is true, in principle, for methods developed in the Earth science community \cite<e.g.,>{Duan_1993, Doherty_2015, Vrugt_2016}.
As a final example, the choice of objective also affects the theoretical limit on data compression.
The basic idea is that we can compress observational data using a scientific model,
then the theoretical compression limit is the storage requirements of our model plus that needed to store the information in the errors.
Minimizing the information in the error, therefore, affords us greater compression
\cite<e.g.>{Rissanen_1978, Grunwald_2000, Grunwald_2007, Weijs_2013, Weijs_2013_zip, Hoge_2018}.

\section{Conclusions}
Many Earth science models are multi-purpose,
so they are difficult or impossible to evaluate except in terms of general utility and information.
In these situations, the information-theoretic paradigm is a natural choice.
Even when there is a specific utility, information theory can help to maximize information about it.
We do not advocate for one objective or another---the choice varies from problem to problem---
only that selecting objectives based on information theory will yield models with less uncertainty,
and we demonstrate one classical approach to doing so.
Despite being somewhat pedagogical, our demonstration provides intuition as to why this choice matters in several contexts
including model selection and evaluation, optimization, uncertainty quantification, and data compression.
It seems reasonable to conclude that such a broad theory might indeed bring a paradigm shift for Earth science \cite{Kumar_2020}.


\appendix

\section{Generalized Likelihood Uncertainty Estimation}
\label{appendix_glue}
Within the context of hydrology, the ideas in this paper closely relate to aspects of
Generalized Likelihood Uncertainty Estimation \cite<GLUE;>[]{Beven_1992}.
We wish to avoid the ``GLUE controversy'' here,
which seems to be, in part, a debate about paradigms \cite<e.g.,>[]{Beven_2013}.
This paper addresses an important gap in the original GLUE framework,
namely, that objective functions (likelihoods) can be evaluated formally using information theory.
Several papers informally explore this idea from a Maximum Likelihood perspective
\cite<e.g.,>{Sorooshian_1980, Beven_2001, Smith_2010, Smith_2015},
but to the best of our knowledge, none of them discuss \citeA{Akaike_1974}.

\section{Relation to Bayesian Inference}
\label{appendix_bayes}
Information criteria, as well as maximum likelihood methods, can be derived from Bayesian theory \cite<>[pp. 301--305]{Burnham_2002_MMI2}.
The practical distinction is that Bayesian methods are more general but also much more computationally demanding.
We argue that information criteria, namely AIC, are pragmatic, easy-to-compute approximations when Bayesian methods are otherwise intractable.
The AIC weights yield a first-order estimate of Bayesian posterior probabilities that any particular model is best given some set of data and models,
and that estimate becomes more and more accurate as sample size increases (Section~\ref{section_aic}).

\section{Adjusted Expectations}
\label{appendix_expectations}
A common response to the information-theoretic paradigm is that modelers select MSE as the objective to ensure
that the model prediction yields the expectation (mean) of the predicted distribution,
as opposed to some other properties like the median.
However, any objective function can yield the expectation by applying the correct adjustment factor.
Under the information-theoretic paradigm,
we instead select the objective (likelihood) to reduce information loss
and then adjust the model's predictions to expectations or any other quantity using an adjustment factor for that likelihood.
This approach will yield a more accurate estimate of the mean (or any other quantity),
because of the invariance property of Maximum Likelihood Estimators:
If $\hat \theta$ is the maximum likelihood estimate (MLE) of $\theta$, then $f(\hat \theta)$ is the MLE $f(\theta)$.
In other words, by reducing a model's uncertainty, we reduce the uncertainty of any estimate derived from that model.

Here is one example of such an adjustment.
A model calibrated to MSLE will predict the median response, but applying an adjustment factor based on $\sigma$ transforms the median into the expectation
\begin{equation}
  \hat y_E = \hat y_M \exp \left( \frac{\sigma^2}{2} \right)
\end{equation}
where $\sigma^2$ is the variance of the errors after transforming the data to natural log scale $\text{Var}(\ln(y) - \ln(\hat y))$ \cite{Limpert_2001}.
These sorts of adjustments are necessary in nearly every application of a model, 
and it is somewhat exceptional that a model calibrated to MSE will predict the expected response.
Keeping with the last example, $\sigma$ also provides the 95-percent prediction intervals or other quantiles
\begin{equation}
  \hat y_{95} = \hat y_M \divideontimes \exp (\sigma 1.96) \text{,}
\end{equation}
or the normal case for which the correction is additive rather than multiplicative,
the prediction interval is
\begin{equation}
  \hat y_{95} = \hat y_E \pm \sigma 1.96 \text{.}
\end{equation}
In this context, $\sigma$ is the only parameter in the normal (and lognormal) likelihood,
so every property of the prediction distribution is some function of $\sigma$.
In most applications, the likelihood is more complicated but the same principle holds.
Therefore, even in an entirely subjective value-based paradigm,
we need a representation of the model's uncertainty to make useful predictions,
and information theory can guide us in selecting one.

\section{NSE Folk Wisdom }
\label{appendix_folkwisdom}
The ``folk wisdom'' among hydrologists is that NSE puts greater emphasis on accuracy at high flows;
whereas, its logged variant (NSLE) puts greater emphasis on low flows.
In some ways that is intuitively obvious, but intuition might obfuscate other effects.
As discussed in \ref{appendix_expectations}, any application of a model should involve some adjustment to the model's prediction,
such as computing prediction intervals or converting predictions to expectations.
For example, when we calibrate to the MSLE, the model's prediction is the median of the conditional distribution,
so we need to apply an adjustment to get the expectation
(failing to do so might be one source for the adage).
For MSLE, that adjustment is straightforward, but this is not always the case.
A common practice is to use the mean of NSE and NSLE as the objective function.
In this case, the prediction is somewhere between the median and the expectation,
and the correction is complicated because these objectives are on different scales (absolute versus relative).

Here, a modeler might apply brute force by using bias adjustment to get the expectation or bootstrapping to get the prediction interval.
As we said previously, some adjustment is \textit{always} necessary,
or at least, we need to convince ourselves that the model's uncertainty is small enough that we can ignore it in our application. 
However, brute-force bias adjustment is in some ways just making up for deficiencies in our original objective---perhaps even negating it---
and in all this complex adjustment, our true objective and uncertainty become somewhat unclear.
We are not against bias adjustment, but it probably works better when paired with a good objective.

\section{Kling--Gupta Efficiency Log-likelihood}
Our experiment has a conspicuous omission:
We failed to derive a log-likelihood equivalent for the Kling-Gupta efficiency \cite<KGE;>{Gupta_2009}.
KGE is a variant of NSE that places additional weight on preserving variance.
It does so by double counting variance in the objective: in both its $r$ and $\alpha$ terms.
Those terms are correlated, so they include redundant information.
For that reason, we expect KGE to perform worse than NSE in an information-theoretic benchmark,
which prefers brevity (compression) over redundancy.

The KGE metric addresses a more practical problem plaguing hydrologic modelers:
their models always \textit{seem} to under-predict variance.
Although the metric is widely used and cited, the original paper stresses a deeper purpose.
\citeA{Gupta_2009} encouraged hydrology to think carefully about its objectives
and demonstrated a general framework for doing so.

\section{Nash--Sutcliffe Efficiency Log-likelihood} \label{appendix_nse}
In this section, we derive the maximum likelihood equivalent log-likelihood for the classic Nash-Sutcliffe efficiency ($R^2$)
and show that is equivalent to that of the normal log-likelihood.
Therefore, optimizing on NSE is equivalent to optimizing on MSE, which is why we use a modified form of NSE in the demonstration (\ref{appendix_loglikelihood}).

The classic definition of NSE is
\begin{equation}
  \text{NSE}(y, \hat{y}) = 1 - \frac{\sum_{i=1}^n (y_i - \hat{y}_i)^2}{\sum_{i=1}^n (y_i - \bar{y})^2} \text{,}
  \label{eq:app-nse}
\end{equation}
where 
$y$ are the observed data,
$\hat{y}$ are the predicted data,
$n$ is the number of observations,
and $\bar{y}$ is the mean of $y$.
In the modified form used in the demonstration, $\bar{y}$ is location-dependent.

We begin with the definition of the mean squared error (MSE)
\begin{equation}
  \text{MSE}(y, \hat{y}) = \frac{1}{n} \sum_{i=1}^n (y_i - \hat{y}_i)^2 \text{.}
  \label{eq:app-mse}
\end{equation}
The equation itself implies nothing about the model's likelihood;
however, whenever we interpret a lower MSE (or higher $R^2$) as ``better'',
we implicitly assume a normal likelihood (consider Anscombe's quartet).
The more the errors deviate from that assumption, the greater the chance that a worse model will yield a lower MSE than a better one.
It is important to understand that the assumption is true of either paradigm:
optimizing a particular ``utility'' on NSE (or MSE) will only minimize risk when the errors are normal
(otherwise, someone can construct a Dutch Book to exploit our ignorance).

This implicit assumption can be understood by examining the equation for the normal log-likelihood ($\ell_2$)
\begin{equation}
  \ell_2(y, \hat{y}) = -\frac{n}{2} \ln(2 \pi \sigma^2) - \frac{1}{2 \sigma^2} \sum_{i=1}^n (y_i - \hat{y}_i)^2
  \label{eq:app-l2}
\end{equation}
where $\sigma^2$ is the variance of the error term ($y - \hat{y}$).
Comparing Equations~\ref{eq:app-mse} and \ref{eq:app-l2}, it is obvious that optimizing either equation will minimize $\sigma^2$,
\begin{equation}
  \sigma^2 = \frac{1}{n} \sum_{i=1}^n (y_i - \hat{y}_i)^2 = \text{MSE} \text{.}
\end{equation}
This is what we mean by $\ell_2$ and MSE are equivalent in the maximum likelihood sense.

Substituting the equation for MSE into the NSE gives
\begin{align}
  \text{NSE}(y, \hat{y}) = 1 - \text{MSE}(y / \sigma_o, \hat{y} / \sigma_o)
  \label{eq:app-mse-substitution}
\end{align}
where $\sigma_o$ is the standard deviation of $y$.
The constant term in equation \ref{eq:app-mse-substitution} does not affect the log-likelihood,
so we can ignore it.
The simplest way to understand that is that we only ever evaluate relative quantities: log-likelihoods, entropies, AICs, etc.
The 1's drop out of the relative NSE, so we can ignore them in the log-likelihood
Similarly, the 1 does not affect the optimization problem except to switch it from maximization to minimization.

Next, substitute the log-likelihoods ($\ell$) into Equation~\ref{eq:app-mse-substitution},
using $\ell_2$ for the MSE.
The log-likelihood is not invariant to transformation,
so we must include the Jacobian term to account for the change of variables.
For log-likelihoods, the change of variables takes the general form of
\begin{equation}
  \ell_Y(y) = \ell_X(v(y)) + \ln | v'(y) |
\end{equation}
where $v$ is the transformation function,
$v'$ is its derivative,
and the second term on the right-hand side is the Jacobian term.
In equation \ref{eq:app-mse-substitution},
the equivalent transformation is $v(y) = y / \sigma_o$ and its derivative is $v'(y) = 1 / \sigma_o$;
therefore, substituting log-likelihoods yields
\begin{equation}
  \ell_{\text{NSE}}(y, \hat{y}) = \ell_2(y / \sigma_o, \hat{y} / \sigma_o) + n \ln | 1 / \sigma_o | \text{.}
  \label{eq:app-l2-substitution}
\end{equation}
Substituting equation \ref{eq:app-l2} for $\ell_2$,
one can verify that the right-hand side is equivalent to $\ell_2(y, \hat{y})$,
\begin{equation}
  \ell_\text{NSE}(y, \hat{y}) = \ell_2(y / \sigma_o, \hat{y} / \sigma_o) + n \ln | 1 / \sigma_o | = \ell_2(y, \hat{y}) \text{.}
  \label{eq:final}
\end{equation}
In other words, the maximum likelihood estimates for the normal log-likelihood ($\ell_2$) and NSE log-likelihood are equivalent.

Here is the step-by-step verification of Equation~\ref{eq:final}
(there are cleaner derivations but we proceed from the previous equation).
First, define the variance of the transformed error $\tau^2$
\begin{align}
  \tau^2 & = \text{Var}(\frac{y-\hat y}{\sigma_o})     \\
         & = \frac{1}{\sigma_o^2} \text{Var}(y-\hat y) \\
         & = \frac{\sigma^2}{\sigma_o^2}
\end{align}
Then proceed with substitutions, beginning with $\ell_2$ in Equation~\ref{eq:final}
\begin{equation}
  \ell_2(y, \hat{y}) = -\frac{n}{2} \ln(2 \pi \tau^2) - \frac{1}{2 \tau^2} \sum_{i=1}^n (v(y_i) - v(\hat{y}_i))^2 + n \ln | 1 / \sigma_o | \text{.}
\end{equation}
then substitute for $\tau$ while following basic rules of logarithms
\begin{align}
  \ell_2(y, \hat{y}) & = -\frac{n}{2} \ln(2 \pi) + n \ln \sigma_o - n\ln \sigma -  \frac{\sigma_o^2}{2 \sigma^2} \sum_{i=1}^n (v(y_i) - v(\hat{y})_i)^2 - n \ln |\sigma_o | \\
                     & = -\frac{n}{2} \ln(2 \pi \sigma^2) -  \frac{\sigma_o^2}{2 \sigma^2} \sum_{i=1}^n (v(y_i) - v(\hat{y})_i)^2
\end{align}
then substituting $v()$
\begin{align}
                     & = -\frac{n}{2} \ln(2 \pi \sigma^2) -  \frac{\sigma_o^2}{2 \sigma^2} \sum_{i=1}^n (y_i/\sigma_o - \hat{y}_i/\sigma_o)^2 \\
                     & = -\frac{n}{2} \ln(2 \pi \sigma^2) -  \frac{\sigma_o^2}{2 \sigma^2 \sigma_o^2} \sum_{i=1}^n (y_i - \hat{y}_i)^2        \\
  \ell_2(y, \hat{y}) & = -\frac{n}{2} \ln(2 \pi \sigma^2) - \frac{1}{2 \sigma^2} \sum_{i=1}^n (y_i - \hat{y}_i)^2 \text{.}
\end{align}
The NSE log-likelihood given in Equation \ref{eq:final} simplifies back to $\ell_2$.

\section{Log-likelihood Derivations} \label{appendix_loglikelihood}    
Log-likelihoods for the other objective functions in the demonstration were derived from $\ell_2$ or $\ell_1$ by changing variables.
The log-likelihood, $\ell$, is a function of $y$, $\hat y$, and $\theta$, denoted as $\ell(y, \hat y, \theta)$,
where $\theta$ are the parameters estimated by maximum likelihood.
For brevity, $\hat y$ and $\theta$ were omitted in the derivations, so the likelihood function is written as
$ell(y)$ or $\ell(v(y))$ when changing variables,
rather than $\ell(v(y), v(\hat y), \theta)$.
The transformation $v()$ applies to both $y$ and $\hat y$,
so the error term in the likelihood changes from $y - \hat y$ to $v(y) - v(\hat y)$.

Using that notation, the log-likelihoods for the other objectives are
(1) mean squared log error (MSLE),
\begin{equation}
  \ell_3 = \ell_2(\ln(y)) + \ln|1/y| \text{,}
\end{equation}
written out as
\begin{equation}
  \ell_3 = -n \ln \sigma  - \frac{n}{2} \ln(2\pi) - \frac{1}{2\sigma^2} \sum_{i=1}^n (\ln(y_i) - \ln(\hat y_i))^2 + \ln|1/y| \text{;}
\end{equation}
(2) normalized mean squared error (NSE; also known as Nash--Sutcliffe efficiency),
\begin{equation}
  \ell_5 = \ell_2(y/\sigma_o) + \ln|1/\sigma_o| \text{;}
\end{equation}
where $\sigma_o$ is a vector containing the standard deviation of the observations at each streamgage location;
(3) mean squared percent error (MSPE),
\begin{equation}
  \ell_6 = \ell_2(y/y) + \ln|-1/y^2| \text{;}
\end{equation}
(4) mean absolute square root error (MARE),
\begin{equation}
  \ell_7 = \ell_1(\sqrt{y}) + \ln|-1/(2\sqrt{y})| \text{.}
\end{equation}
In each, the second term is the derivative of the transformation from the change of variables;
so in $\ell_3$, $1/y$ is the derivative of $\ln(y)$.
The derivative term could be simplified to $-\ln(y)$,
but we left it to make the derivation clearer.
Log likelihoods for MALE, and ZMALE are represented by substituting $\ell_1$ for $\ell_2$.
Finally, the log-likelihood of the uniform error (U) is
\begin{equation}
  \ell_8 = -n \ln( \text{max}(|y - \hat y|) ) \text{,}
\end{equation}
which is an objective that minimizes the maximum error. 

Note the distinction between NSE and MSE:
In NSE, the errors are normalized by dividing them by the variance of the observations at that location $\sigma_o$; whereas
in MSE, the errors are left in their original units.
By that definition, their log-likelihoods will be equivalent when measured at a single location but will differ when measured over multiple locations with different variances
(\ref{appendix_nse} proves the former claim for large sample sizes).

\section{The Problem with Some Likelihoods} \label{appendix_limitations}
S.V. Weijs points out an important flaw with using AIC to compare zero-inflated and non-zero-inflated likelihoods,
which also happens to be an instance of a much broader and, to our knowledge, unresolved problem in information theory.

Using a change of variables (Equation~\ref{eq.change-of-variables}) ensures that rescaling the data
will not alter the likelihood (Otherwise, we could alter our results by changing units!).
However, this ``invariance'' only holds so far.
If we assume different starting units, we will observe a different magnitude for the likelihood.
In other words, the likelihood is invariant but not ``coordinate independent.''
This is not a problem for univariate likelihoods,
because the likelihood ratios and Akaike weights are still invariant,
and that is sufficient.

Coordinate dependence only becomes a problem when we consider a multivariate likelihood (objective),
such as the joint likelihood of streamflow and stream temperature
or the flow-state and flow-magnitude components of a zero-inflated likelihood ($\ell_0$ and $\ell_3$ in Equation~\ref{eq.zero_inflated_likelihood}).
Here, assuming different starting units will alter the likelihood ratios and Akaike weights.
Therefore, multivariate likelihoods will involve some subjective decision-making,
but theory still goes a long way.
In our demonstration, we are confident that a log or square-root transformation will reduce information loss (Table~\ref{table1}).
If we decide to use log, then we recommend trying zero-inflation or other strategies for handling the no-flow state.
Many readers are already aware of this problem from
multi-objective calibration where it produces a ``Pareto front of optimality.''
We will not attempt to solve the multivariate problem here, but we believe there are ways forward.

%
%

\section*{Open Research Section}
All code used in the demonstration is available as a Jupyter notebook at \url{https://doi.org/10.5066/P983Q1D2}.
The streamflow data are from \citeA{Russell_2020} and are available at \url{https://doi.org/10.5066/P9XT4WSP}.
That data release includes predictions from several statistical models;
the demonstration uses the nearest neighbor drainage area ratio (NNDAR).

\acknowledgments
The authors thank Hoshin V. Gupta for his encouragement and thoughtful discussion about information theory while waiting in an airport after HydroML 2022,
which provided the inspiration for this paper.
The authors also thank Jasper A. Vrugt, Steven V. Weijs, Uwe Ehret, and Thorsten Wagener for their substantial efforts to improve this manu\-script.
In particular, JAV assisted in the NSE log-likelihood derivation in \ref{appendix_nse}.
Funding for this research was provided by the Hydro-terrestrial Earth Systems Testbed (HyTEST) project of the U.S. Geological Survey Integrated Water Prediction program.

%
%

\bibliography{paper.bib}

%
%
%
%
%

\end{document}